\documentclass[11pt]{article}

\PassOptionsToPackage{table}{xcolor}

\usepackage[preprint]{acl}

\usepackage{times}
\usepackage{latexsym}

\usepackage[T1]{fontenc}

\usepackage[utf8]{inputenc}

\usepackage{microtype}

\usepackage{inconsolata}

\usepackage{graphicx}
\usepackage{algorithm}
\usepackage{algorithmic}
\usepackage{bm}
\usepackage{amsmath}
\usepackage{amssymb}
\usepackage{tcolorbox}
\usepackage{fvextra}
\DefineVerbatimEnvironment{MyVerbatim}{Verbatim}{
  breaklines=true,
  breaksymbolleft={},   
  breaksymbolright={}   
}
\usepackage{hyperref}
\usepackage{url}
\usepackage{booktabs}

\title{GraphER: An Efficient Graph-Based Enrichment and Reranking Method for Retrieval-Augmented Generation}

\author{
  \textbf{Ruizhong Miao},
  \textbf{Yuying Wang},
  \textbf{Rongguang Wang},
  \textbf{Chenyang Li},
\\
  \textbf{Tao Sheng},
  \textbf{Sujith Ravi},
  \textbf{Dan Roth}
\\
\\
  Oracle AI
}

\begin{document}
\maketitle
\begin{abstract}

Semantic search in retrieval-augmented generation (RAG) systems is often insufficient for complex information needs, particularly when relevant evidence is scattered across multiple sources, because it may fail to retrieve the \textit{complete} set of evidence. Existing approaches to addressing this problem either rely on iterative agentic retrieval, which can be computationally inefficient, or maintain additional structures such as knowledge graphs, which introduce storage and maintenance overhead. In this paper, we propose \textbf{GraphER}, a graph-based enrichment and reranking framework that (1) leverages the organizational structure of data to capture proximity relationships beyond semantic similarity, (2) constructs a graph at query time based on these proximities, and (3) applies graph-based ranking to surface the top candidate documents. Experiments across table retrieval, multi-hop retrieval, and long-document retrieval benchmarks demonstrate consistent improvements in terms of \textit{retrieval completeness}. Additionally, GraphER requires no additional graph infrastructure and integrates seamlessly with standard vector stores. The framework is retriever-agnostic, supports multiple forms of proximity, and introduces minimal query-time latency.

\end{abstract}

\section{Introduction}
\label{sec:Introduction}

In retrieval-augmented generation (RAG; \citealp{lewis2020retrieval,yasunaga2023retrieval}) systems, incomplete retrieval is often a primary bottleneck to answer quality. Modern neural retrievers estimate query-document relevance based on semantic similarity. For example, embedding-based bi-encoders and cross-encoders map queries and documents into vector representations and rank documents according to their proximity in the embedding space. Despite their scalability and effectiveness, these models often struggle with complex queries that require integrating information from multiple sources. This limitation arises mainly due to (1) they are point-wise rankers that evaluate each query-document pair in isolation without considering the relationships among candidate documents; and (2) semantic proximity alone may not fully capture the organizational structure of the data within the retrieval corpus.

\begin{figure}[t]
  \begin{center}
    \centerline{\includegraphics[width=\columnwidth]{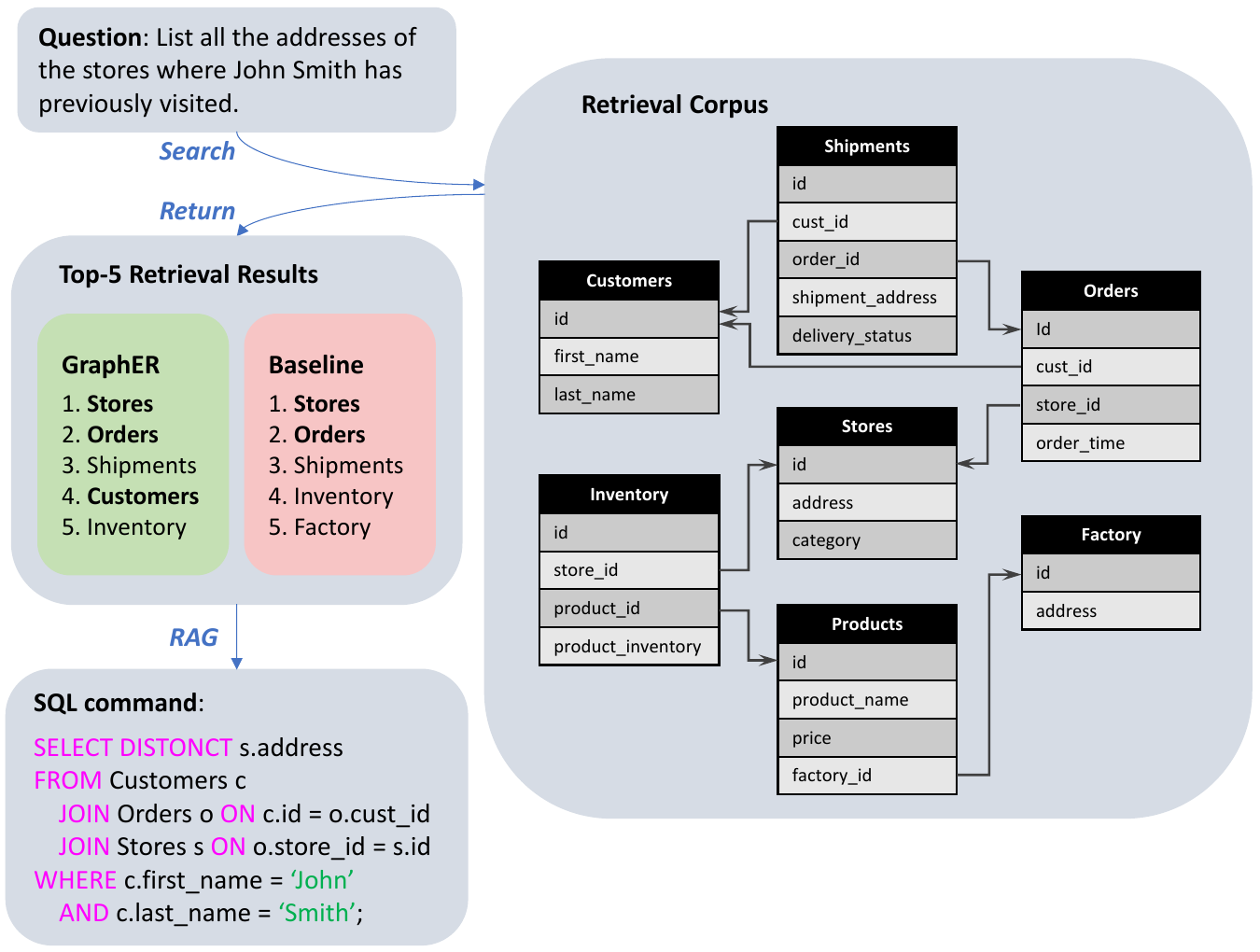}}
    \caption{
Example of RAG for SQL generation: Semantic search misses the Customers table in the top-5 results, whereas GraphER reranks the candidates to retrieve all relevant tables.
    }
    \label{fig:SQL_Retrieval_Example}
  \end{center}
\end{figure}

Consider the table retrieval example in Figure \ref{fig:SQL_Retrieval_Example}. To generate the correct SQL query, the retriever must return all relevant tables: \textit{Customers}, \textit{Orders}, and \textit{Stores}. Semantic search ranks \textit{Orders} and \textit{Stores} highly because of their semantic proximity to to the question, but fails to return \textit{Customers} in the top-5 results. Although \textit{Customers} is required to generate the correct SQL query, its relevance stems primarily from its foreign-key relationship with \textit{Orders} rather than its semantic similarity to the question.

These observations motivate retrieval methods that consider not only semantic proximity between query-document pairs but also relationships among candidate documents. Prior work has explored several approaches to improve retrieval completeness in RAG. Agentic retrieval methods use LLM-guided iterative search to decompose queries and gather evidence across multiple retrieval steps \citep{yao2023react,khot2023decomposed,zhou2023leasttomost,trivedi2023interleaving,asai2024self}. The agentic approach is well-suited for retrieving information from external, open-ended environments (e.g., web search), where the data organization is largely unknown and users lack control over the indexing process. However, for private or enterprise databases, where users do have control over the organization of the available data, these methods can incur unnecessary latency and cost due to repeated retrieval and reasoning. Alignment-oriented approaches \citep{chen2025can} attempt to jointly reason over content and structure, but they also rely on computationally expensive online LLM calls and constrained decoding, which limit their practicality in many production settings.

Another class of methods augments retrieval with auxiliary data structures, such as knowledge graphs or hierarchical indexes, to capture relationships beyond semantic proximity \citep{sun2024thinkongraph,luo2024reasoning,jimenez2024hipporag,gutierrez2025from,sarthi2024raptor,sourati2026lad}. These approaches require maintaining additional infrastructure alongside the base retriever. While effective in their respective settings, such infrastructure introduces computational and storage overhead, limiting their applicability in general-purpose RAG systems.

To address these limitations, we introduce \textbf{GraphER}, a \textbf{Graph}-based \textbf{E}nrichment and \textbf{R}eranking method for RAG systems. GraphER is capable of incorporating various forms of proximity relationships beyond semantic proximity. Its objective is to improve the completeness of the top-K results by exploiting these proximity relationships among candidate objects. GraphER is designed to be a standalone and computationally efficient reranking stage that is compatible with other upstream and downstream reranking stages. Therefore, GraphER can be combined with more computationally intensive semantic rerankers, such as cross-encoder rerankers.

From an efficiency perspective, unlike knowledge graph-based approaches, GraphER requires no additional infrastructure beyond a base retriever. Additionally, any potential LLM calls in GraphER are confined to the offline phase, while online retrieval is LLM-free and incurs minimal query-time latency. These advantages make GraphER easy to integrate into most production RAG systems.

In summary, our contributions are as follows:
\begin{itemize}
    \item We introduce \textbf{GraphER}, a graph-based enrichment and reranking framework that improves retrieval completeness in RAG systems.
    \item We propose \textbf{Graph Cohesive Smoothing (GCS)}, a lightweight graph-based ranking algorithm that mitigates hub-node bias.
    \item We demonstrate that GraphER naturally supports graph neural network rerankers.
    \item We conduct extensive experiments on three distinct types of retrieval tasks to validate the effectiveness of GraphER.
\end{itemize}

\begin{figure*}[!t]
  \begin{center}
    \centerline{\includegraphics[width=\textwidth]{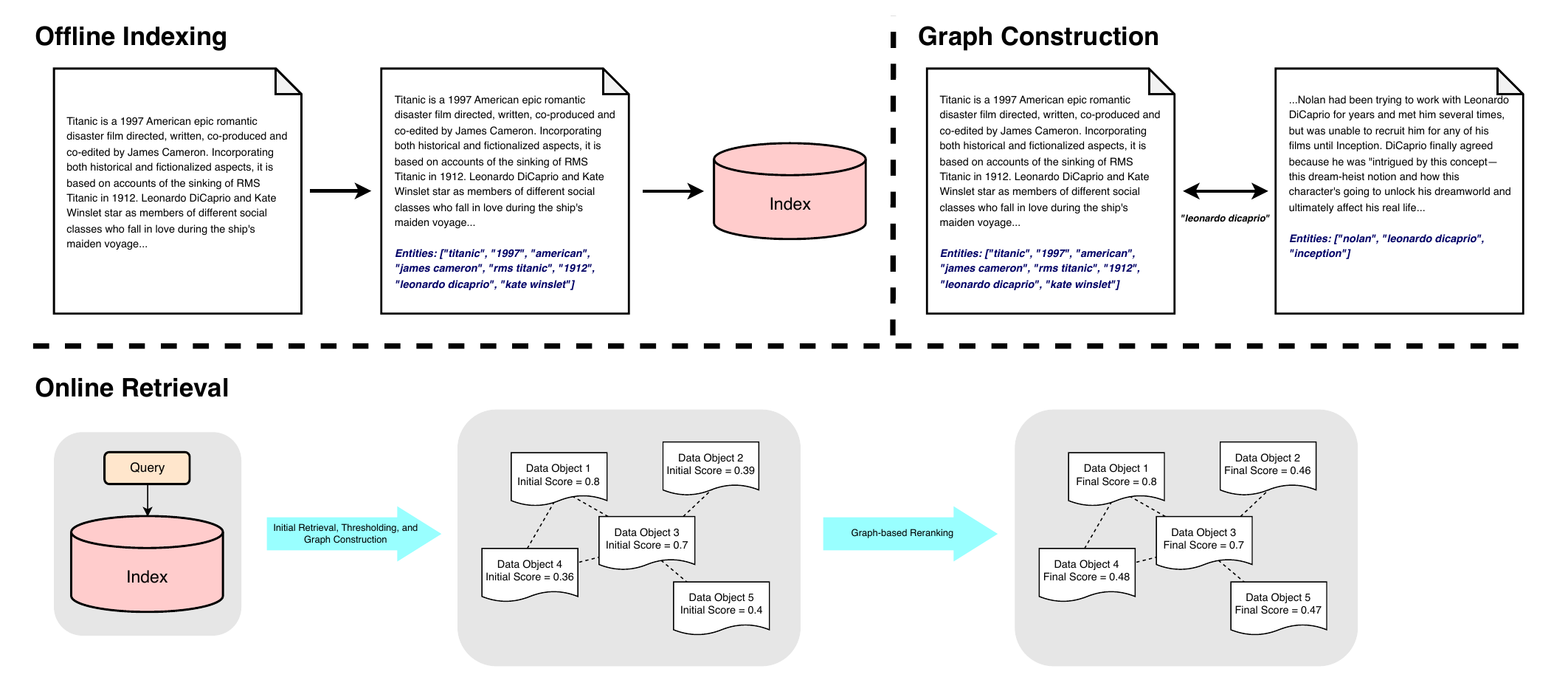}}
    \caption{
During offline indexing, each data object is enriched with metadata that informs edge relationships and then indexed using standard retrieval methods. During online retrieval, a base retriever retrieves candidate objects and assigns initial scores, after which a graph is constructed from the candidates based on the enrichments. A graph-based ranking algorithm is then applied to produce the final relevance scores.
    }
    \label{fig:graph_offline_online}
  \end{center}
\end{figure*}

\section{Methodology}
\label{sec:Methodology}

GraphER consists of two main components: (1) an offline enrichment process that incorporates graph information during indexing, and (2) an online graph-based reranking process that leverages the enriched graph information to refine the initial search results. Figure \ref{fig:graph_offline_online} gives an overview of the GraphER methodology. More details on these components are provided below.

\subsection{Offline indexing enriched with graph information}\label{sec:graph_enrichment}

During offline indexing, we enrich each data object\footnote{The term data object here can refer to either a text passage or a table that has been serialized into text format.} with metadata that encode potential graph relationships to other data objects. In this work, we consider three types of proximity relationships:

\paragraph{Structural proximity} Metadata are derived from predefined relationships such as foreign-key links between tables or hyperlinks between webpages.

\paragraph{Conceptual proximity} We extract named entities from each data object using an instruction-tuned LLM and store them as metadata. Two objects are considered conceptually close if they share one or more named entities.

\paragraph{Contextual proximity} We record document and chunk identifiers to link adjacent chunks from the same document.

Using these metadata, a graph edge can be created between two data objects when they satisfy the corresponding proximity criterion. Since each object is enriched independently, the process requires only a single pass through the corpus and has linear complexity with respect to the index size.

\subsection{Online graph construction and graph-based reranking}

The online retrieval process consists of four steps: candidate retrieval, graph construction, graph-based reranking, and outputting the final ranked list. The candidate retrieval step is a standard procedure in RAG systems: the base retriever retrieves the top-$n$ candidate objects and assigns an initial relevance score to each of them.

\paragraph{Graph construction}
Given the retrieved candidates, we construct a graph in which each node represents a candidate data object. An edge is added between two nodes if their metadata indicate a proximity relationship, as described in Section \ref{sec:graph_enrichment}. For example, two passages enriched with named entities are connected if they share at least one named entity.

Note that the graph here differs from a knowledge graph in two key respects. First, graph construction occurs only for candidate objects during query time, and no knowledge graph is maintained in the index. This enables GraphER to integrate seamlessly with standard vector stores. Second, unlike a knowledge graph, where nodes represent entities and edges represent relationships, in GraphER a node represents a data object, which is typically a text string containing multiple entities, and the edges are unlabeled.

{\setlength{\textfloatsep}{10pt plus 1pt minus 2pt}
\begin{algorithm}[t]
  \caption{Graph Cohesive Smoothing}
  \label{alg:gcs}
  \begin{algorithmic}
    \STATE {\bfseries Input:} Adjacency matrix $\bm{A}\in\mathbb{R}^{n\times n}$ with $\bm{A}_{ij} \ge 0$, seed score vector $\bm{s} \in \mathbb{R}^{n}$, damping factor $\alpha \in (0,1)$, tolerance $\epsilon>0$.
    \STATE {\bfseries Output:} Relevance score vector $\bm{p} \in \mathbb{R}^{n}$
    \STATE {\bfseries Definition:} Let $\bm{W}$ be the row-normalized transition matrix, where
    \[
       \bm{W}_{ij} = 
       \begin{cases}
         \frac{\bm{A}_{ij}}{\sum_{k=1}^n \bm{A}_{ik} } & \text{if } \bm{A}_{ij} > 0 \\
         0 & \text{otherwise}.
       \end{cases}
    \]
    \STATE \textbf{initialize} $\bm{p}^{(0)} = \bm{s}$, $t = 0$.
    \REPEAT
      \STATE $\bm{p}^{(t+1)} = \alpha \bm{s} + (1 - \alpha)\,\bm{W}\,\bm{p}^{(t)}$
      \STATE Compute residual $\delta = \| \bm{p}^{(t+1)} - \bm{p}^{(t)} \|_1$
      \STATE $t = t + 1$
    \UNTIL{$\delta < \epsilon$}
    \STATE \textbf{return} $\max(\bm{p}^{(t)}, \bm{s})$ (element-wise maximum)
  \end{algorithmic}
\end{algorithm}
}

\paragraph{Graph-based ranking} We then run graph-based ranking algorithms on the constructed graph. Personalized PageRank (PPR) has been used in various RAG systems \citep{jimenez2024hipporag,gutierrez2025from,xu2026from} to rank candidate objects. As a comparative graph ranking algorithm, we also include PPR in our experiments. However, PPR has a tendency to favor hub nodes, i.e., nodes with a large number of connections to other nodes. This occurs even when both the hub node itself and its neighbors have low initial scores. This is due to PPR's random walk mechanism, which causes the hub node to be frequently visited and accumulate probability flows from its numerous neighbors. As a result, the hub node tends to be ranked highly, regardless of its actual relevance to the query.

\setlength{\tabcolsep}{2.5pt}
\begin{table*}[h]
  \begin{center}
      \resizebox{\linewidth}{!}{
        \begin{tabular}{lccccccccccc}
          \toprule
           & Spider1\_train & Spider1\_dev & Spider1\_test & Bird\_train & Bird\_dev & Beaver\_dev & HotpotQA & 2WikiMultihopQA & MuSiQue & BEIR-NQ \\
          \midrule
          \# queries                                  & 8650 & 1032 & 2147 & 9398 & 1534 & 209 & 2000   & 2000  & 2417   & 3452  \\
          \# queries with $>$1 rel. object & 3658 & 378   & 787   & 7679 & 1218 & 206 & 2000   & 2000  & 2417   & 666  \\
          \# data objects                           & 876   & 876   & 1056 & 522   & 75    & 463 & 19323 & 12601 & 21100 & 117683  \\
          \bottomrule
        \end{tabular}}
  \end{center}
  \caption{Dataset statistics.}
  \label{tab:dataset_statistics}
\end{table*}

To address this limitation of PPR, we draw on the principle of graph cohesion \citep{li2019prediction,le2022linear}, which assumes that connected nodes tend to exhibit similar characteristics. Based on this principle, we propose an iterative ranking algorithm, termed Graph Cohesive Smoothing (GCS). The algorithm is described in Algorithm \ref{alg:gcs}.

Compared with PPR, GCS uses row normalization, which smooths scores across neighboring nodes rather than propagating probability mass through random walks. Furthermore, the final element-wise maximum operation ensures that each node's score is at least as high as its initial score, which prevents highly relevant nodes from being penalized by averaging with lower-scoring neighbors.

\paragraph{Graph attention network ranker} The ranking of candidate data objects can be viewed as a node-level prediction task. When the base retriever involves an embedding model, we also train a graph attention network (GAT) \citep{velivckovic2018graph} to leverage the nodal features and the graph structure defined on the candidate data objects. Specifically, the GAT architecture used in our experiments consists of five GATv2 layers \citep{brody2022how}, followed by two fully connected layers. For each graph node, the GAT takes as input a feature vector formed by concatenating the GCS score, the query embedding, and the data object embedding, and outputs a scalar value that serves as the final ranking score for that node.

\section{Experiments}
\label{sec:Experiments}

\subsection{Experimental Setup}

\paragraph{Datasets} We evaluate GraphER on three task types, using a number of different datasets for each. The three task types are: (1) \textit{table retrieval}, for which we use the Spider 1.0 \citep{yu2018spider}, Bird \citep{li2023can}, and Beaver \citep{chen2025beaver} datasets; (2) \textit{multi-hop QA retrieval}, for which we use the HotpotQA \citep{yang2018hotpotqa}, 2WikiMultihopQA \citep{ho2020constructing}, and MuSiQue \citep{trivedi2022musique} datasets. For HotpotQA and 2WikiMultihopQA, we randomly sampled 2000 queries from their dev branches and included the corresponding passages of the sampled queries as the retrieval corpus. For MuSiQue, we use the entire \texttt{musique\_ans\_v1.0\_dev} branch; (3) \textit{chunked documents retrieval}, for which we use the BEIR-NQ \citep{thakur2021beir} dataset. For this dataset, we only keep documents that contain at least one chunk identified as a relevant object for one or more queries.

Each of these three task types corresponds to a specific type of proximity described in Section \ref{sec:graph_enrichment}. Specifically, in addition to semantic proximity, we consider structural proximity for table retrieval tasks, conceptual proximity for multi-hop QA retrieval tasks, and contextual proximity for chunked document retrieval tasks.

In each of the datasets considered, a significant portion of the queries have more than one relevant object for retrieval. The detailed statistics of these datasets are listed in Table \ref{tab:dataset_statistics}.

\begin{table*}[!t]
  \begin{center}
    \begin{scriptsize}
        \resizebox{\linewidth}{!}{
        \begin{tabular}{llllll}
          \toprule
          & \multicolumn{5}{c}{\textbf{Dataset}} \\
          \cmidrule(lr){2-6}
          \textbf{Retriever} & Spider1\_train & Spider1\_dev & Spider1\_test & Bird\_dev & Beaver\_dev \\
          \midrule
          \multicolumn{6}{c}{PR@5 (\%)} \\
          \midrule
          Llama-Embed-Nemotron-8B+BM25                                    & 23.7 (42.5) & 37.3 (57.3) & 22.2 (41.8) & 21.8 (30.8) & 14.1 (14.8) \\
          Llama-Embed-Nemotron-8B+BM25 $+$ GraphER-PPR    & \textbf{29.7} (\underline{44.4}) & 44.7 (58.8) & 26.9 (41.6) & \underline{33.0} (\underline{40.3}) & 11.7 (12.4) \\
          Llama-Embed-Nemotron-8B+BM25 $+$ GraphER-GCS    & \underline{29.4} (\textbf{44.9}) & \underline{47.6} (\underline{61.0}) & \underline{29.2} (\underline{44.6}) & 29.4 (36.6) & \textbf{17.0} (\textbf{17.7}) \\
          Llama-Embed-Nemotron-8B+BM25 $+$ GraphER-GAT    & \textit{N/A}    & \textbf{49.7} (\textbf{62.5}) & \textbf{40.9} (\textbf{54.4}) & \textbf{36.9} (\textbf{43.2}) & \underline{15.0} (\underline{15.8}) \\
          \rowcolor{gray!20}
          \textit{GraphER-GAT's improvement}                      & \textit{N/A} & $+12.4$ ($+5.2$) & $+18.7$ ($+12.5$) & $+15.2$ ($+12.3$) & $+1.0$ ($+1.0$) \\
          \midrule
          Cohere-Embed-V4+BM25                                                   & 46.6 (71.0) & 75.9 (89.2) & 50.4 (75.3) & 74.2 (79.3) & 11.2 (12.4) \\
          Cohere-Embed-V4+BM25 $+$ GraphER-PPR                   & \textbf{55.0} (\underline{73.1}) & \textbf{85.2} (\textbf{92.4}) & \underline{63.7} (78.9) & \textbf{86.2} (\textbf{88.4}) & 10.7 (12.0) \\
          Cohere-Embed-V4+BM25 $+$ GraphER-GCS                   & \underline{52.9} (\textbf{73.7}) & \underline{80.4} (\underline{91.1}) & 63.0 (\underline{80.3}) & 80.0 (83.9) & \underline{12.6} (\textbf{13.9}) \\
          Cohere-Embed-V4+BM25 $+$ GraphER-GAT                   & \textit{N/A} & 80.2 (90.8) & \textbf{65.7} (\textbf{81.8}) & \underline{83.4} (\underline{86.7}) & \textbf{13.1} (\textbf{13.9}) \\
          \rowcolor{gray!20}
          \textit{GraphER-GAT's improvement}                      & \textit{N/A} & $+4.2$ ($+1.6$) & $+15.2$ ($+6.5$) & $+9.2$ ($+7.4$) & $+1.9$ ($+1.4$) \\
          \midrule
          \multicolumn{6}{c}{PR@10 (\%)} \\
          \midrule
          Llama-Embed-Nemotron-8B+BM25                                    & 38.8 (57.5) & 58.7 (73.6) & 38.5 (59.6) & 70.4 (75.7) & \underline{30.6} (\underline{31.1}) \\
          Llama-Embed-Nemotron-8B+BM25 $+$ GraphER-PPR    & \underline{44.3} (\underline{59.1}) & 66.7 (75.7) & 46.1 (61.1) & \textbf{78.2} (\textbf{81.4}) & 20.9 (21.5) \\
          Llama-Embed-Nemotron-8B+BM25 $+$ GraphER-GCS    & \textbf{45.1} (\textbf{59.9}) & \underline{69.0} (\underline{77.4}) & \underline{50.4} (\underline{64.2}) & 72.8 (77.0) & \textbf{31.1} (\textbf{31.6}) \\
          Llama-Embed-Nemotron-8B+BM25 $+$ GraphER-GAT    & \textit{N/A} & \textbf{70.1} (\textbf{78.5}) & \textbf{59.1} (\textbf{71.0}) & \underline{77.0} (\underline{80.4}) & 29.6 (30.1) \\
          \rowcolor{gray!20}
          \textit{GraphER-GAT's improvement}                      & \textit{N/A} & $+11.4$ ($+4.8$) & $+20.6$ ($+11.4$) & $+6.6$ ($+4.6$) & $-1.0$ ($-1.0$) \\
          \midrule
          Cohere-Embed-V4+BM25                                                   & 60.3 (80.4) & 86.0 (93.8) & 65.2 (84.8) & 89.3 (91.5) & 19.9 (21.1) \\
          Cohere-Embed-V4+BM25 $+$ GraphER-PPR                   & \textbf{67.4} (\underline{82.6}) & 94.2 (97.3) & \underline{75.7} (88.2) & \underline{95.9} (\underline{96.6}) & 19.4 (20.6) \\
          Cohere-Embed-V4+BM25 $+$ GraphER-GCS                   & \underline{66.5} (\textbf{83.0}) & \underline{95.2} (\underline{97.5}) & 75.2 (\underline{88.4}) & 95.5 (96.3) & \underline{21.4} (\underline{22.5}) \\
          Cohere-Embed-V4+BM25 $+$ GraphER-GAT                   & \textit{N/A} & \textbf{96.3} (\textbf{97.8}) & \textbf{77.4} (\textbf{89.7}) & \textbf{96.6} (\textbf{97.3}) & \textbf{23.3} (\textbf{24.4}) \\
          \rowcolor{gray!20}
          \textit{GraphER-GAT's improvement}                      & \textit{N/A} & $+10.3$ ($+4.1$) & $+12.2$ ($+4.9$) & $+7.3$ ($+5.8$) & $+3.4$ ($+3.3$) \\
          \bottomrule
        \end{tabular}
        }
    \end{scriptsize}
  \end{center}
  \caption{Retrieval performance on table retrieval benchmarks. We highlight the \textbf{best} and \underline{second-best} results within each dataset-base retriever setting. Numbers in parentheses denote PR@K over all queries; numbers outside parentheses denote PR@K for queries with more than one relevant object.}
  \label{tab:table_retrieval_eval}
\end{table*}

\paragraph{Base retriever} We use a hybrid search approach as the base retriever, which combines semantic embedding and keyword search. For embedding models, we use \textbf{Llama-Embed-Nemotron-8B} \citep{babakhin2025llama} and \textbf{Cohere-Embed-4} \citep{cohere2025embed}. For the BEIR-NQ dataset in particular, we use \textbf{Llama-Embed-Nemotron-8B} and \textbf{Multilingual-E5-large} \citep{wang2024multilingual}. For each of these embedding models, we combine it with the \textbf{BM25} algorithm \citep{robertson2009probabilistic,trotman2014improvements} by calculating a weighted average of the normalized BM25 score and the cosine similarity between the query and document embeddings. The weights used are 0.3 for BM25 and 0.7 for cosine similarity, respectively.

Our evaluation aims to isolate the contribution of GraphER beyond semantic retrieval. Therefore, we evaluate GraphER on top of strong semantic retrieval backbones rather than more sophisticated semantic rerankers. This allows us to measure the incremental benefit of the proposed methodology beyond the underlying semantic retriever.

In all experiments, the base retriever retrieves 200 candidate objects, or all available objects if there are fewer than 200. A graph is then constructed from these candidate objects, and GraphER's reranking algorithm is applied.

\paragraph{Metrics} To evaluate retrieval completeness, we use \textit{perfect} recall@K (PR@K). For a given question, PR@K is 1 if \textit{all} relevant objects are included in the top-K search results, and 0 otherwise. In contrast to regular recall@K, which measures the proportion of relevant objects retrieved, PR@K is particularly important for questions that require information from multiple sources, as failing to retrieve even a single relevant object can result in incorrect answers.

For multi-hop QA retrieval tasks, we also use the top-10 retrieved data objects to generate answers for both GraphER-GAT and the baseline using GPT-5 \citep{openai2025gpt}. The generated answers are evaluated by prompting GPT-5 to compare them against the ground-truth answers. When available, we report the percentages of queries answered correctly, as judged by GPT-5. The prompt templates for answer generation and evaluation are provided in Appendices \ref{appendix:prompt_rag} and \ref{appendix:prompt_rag_eval}.

\begin{table*}[!t]
  \begin{center}
    \begin{scriptsize}
        \resizebox{\linewidth}{!}{
        \begin{tabular}{llll}
          \toprule
          & \multicolumn{3}{c}{\textbf{Dataset}} \\
          \cmidrule(lr){2-4}
          \textbf{Retriever} & HotpotQA & 2WikiMultihopQA & MuSiQue \\
          \midrule
          \multicolumn{4}{c}{PR@5 (\%)} \\
          \midrule
          Llama-Embed-Nemotron-8B+BM25                                    & \underline{78.5} & 42.6 & 24.2 \\
          Llama-Embed-Nemotron-8B+BM25 $+$ GraphER-PPR    & 72.9 & 42.1 & 17.3 \\
          Llama-Embed-Nemotron-8B+BM25 $+$ GraphER-GCS    & \textbf{78.8} & \underline{43.8} & \underline{25.4} \\
          Llama-Embed-Nemotron-8B+BM25 $+$ GraphER-GAT    & 78.0 & \textbf{44.1} & \textbf{25.6} \\
          \rowcolor{gray!20}
          \textit{GraphER-GAT's improvement}                      & $-0.5$ & $+1.5$ & $+1.4$ \\
          \midrule
          Cohere-Embed-V4+BM25                                                   & 74.2 & 42.0 & 23.0 \\
          Cohere-Embed-V4+BM25 $+$ GraphER-PPR                   & 71.6 & \textbf{45.7} & 22.1 \\
          Cohere-Embed-V4+BM25 $+$ GraphER-GCS                   & \underline{74.4} & 43.3 & \underline{24.0} \\
          Cohere-Embed-V4+BM25 $+$ GraphER-GAT                   & \textbf{74.7} & \underline{43.8} & \textbf{24.5} \\
          \rowcolor{gray!20}
          \textit{GraphER-GAT's improvement}                      & $+0.4$ & $+1.8$ & $+1.5$ \\
          \midrule
          \multicolumn{4}{c}{PR@10 (\%)} \\
          \midrule
          Llama-Embed-Nemotron-8B+BM25                                    & 88.2 (QA Acc. = 84.2) & 49.4 (QA Acc. = 43.1) & 35.5 (QA Acc. = 44.1) \\
          Llama-Embed-Nemotron-8B+BM25 $+$ GraphER-PPR    & 84.5 & 49.6 & 27.7 \\
          Llama-Embed-Nemotron-8B+BM25 $+$ GraphER-GCS    & \underline{88.7} & \underline{51.1} & \underline{36.9} \\
          Llama-Embed-Nemotron-8B+BM25 $+$ GraphER-GAT    & \textbf{88.9} (QA Acc. = 84.9) & \textbf{53.0} (QA Acc. = 45.4) & \textbf{37.4} (QA Acc. = 45.2) \\
          \rowcolor{gray!20}
          \textit{GraphER-GAT's improvement}                      & $+0.6$ (QA Acc. $+0.7$) & $+3.6$ (QA Acc. $+2.3$) & $+1.9$ (QA Acc. $+1.0$) \\
          \midrule
          Cohere-Embed-V4+BM25                                                   & 85.5 (QA Acc. = 82.2) & 48.3 (QA Acc. = 43.5) & 34.1 (QA Acc. = 42.2) \\
          Cohere-Embed-V4+BM25 $+$ GraphER-PPR                   & 85.8 & \textbf{52.8} & 34.6 \\
          Cohere-Embed-V4+BM25 $+$ GraphER-GCS                   & \underline{86.4} & 50.1 & \underline{35.5} \\
          Cohere-Embed-V4+BM25 $+$ GraphER-GAT                   & \textbf{87.0} (QA Acc. = 83.7) & \underline{50.7} (QA Acc. = 44.9) & \textbf{36.3} (QA Acc. = 43.5) \\
          \rowcolor{gray!20}
          \textit{GraphER-GAT's improvement}                      & $+1.4$ (QA Acc. $+1.5$) & $+2.4$ (QA Acc. $+1.4$) & $+2.2$ (QA Acc. $+1.3$) \\
          \bottomrule
        \end{tabular}
        }
    \end{scriptsize}
  \end{center}
  \caption{Retrieval performance on multi-hop QA benchmarks. Numbers in parentheses denote the percentages of queries answered correctly, as evaluated by GPT-5.}
  \label{tab:multihop_retrieval_eval}
\end{table*}

\paragraph{Hyper-Parameter Tuning and Model Training Protocols} Both PPR and GCS have one hyper-parameter, namely the damping factor $\alpha$. For each base retriever setting, specifically Llama-Embed-Nemotron-8B+BM25 and Cohere-Embed-4+BM25, we perform a grid search over $\alpha$, evaluating candidate values from 0.1 to 0.9 in increments of 0.1 on the \texttt{Bird\_train} dataset. The best-performing value of $\alpha$ on this dataset is then selected and fixed for the remaining datasets. For the Multilingual-E5-large+BM25 setting, which is only applied to the BEIR-NQ dataset, we use the same value of $\alpha$ as in Llama-Embed-Nemotron-8B+BM25.

In both PPR and GCS, edge weights of the adjacency matrix $\bm{A}$ are user-specified hyper-parameters that are predetermined and not subject to tuning. Throughout our experiments, we use the same edge weights within each task type. Specifically, for table retrieval datasets, we assign an edge weight of 1 between two tables if they are linked through a foreign-key relationship, and 0 otherwise. Similarly, for chunked document retrieval, we set the edge weight between two chunks to 1 if they are adjacent chunks from the same un-chunked document, and 0 otherwise. For multi-hop QA retrieval, we compute the edge weight $\bm{A}_{ij}$ as the number of unique named entities shared by objects $i$ and $j$, divided by the total number of unique named entities in object $j$. This normalization helps control the influence of passages that contain a large number of entities by diluting their emphasis on any single entity in proportion to the total number of entities they contain. The extraction of named entities for each data object is done by prompting GPT-4o \citep{openai2024gpt}. The prompt template is provided in Appendix \ref{appendix:prompt_ner}.

\begin{table}[!t]
  \begin{center}
    \begin{scriptsize}
    	\resizebox{\linewidth}{!}{
        \begin{tabular}{llll}
          \toprule
          & \textbf{Dataset} \\
          \cmidrule(lr){2-2}
          \textbf{Retriever} & BEIR-NQ \\
          \midrule
          \multicolumn{2}{c}{PR@5 (\%)} \\
          \midrule
          Llama-Embed-Nemotron-8B+BM25                                    & 50.9 (77.4) \\
          Llama-Embed-Nemotron-8B+BM25 $+$ GraphER-PPR    & \underline{53.3} (76.8) \\
          Llama-Embed-Nemotron-8B+BM25 $+$ GraphER-GCS    & 52.3 (\underline{78.1}) \\
          Llama-Embed-Nemotron-8B+BM25 $+$ GraphER-GAT    & \textbf{54.4} (\textbf{78.2}) \\
          \rowcolor{gray!20}
          \textit{GraphER-GAT's improvement}                      & $+3.5$ ($+0.8$) \\
          \midrule
          Multilingual-E5-large+BM25                                                          & \underline{10.7} (\underline{25.2}) \\
          Multilingual-E5-large+BM25 $+$ GraphER-PPR                          & 7.2 (20.9) \\
          Multilingual-E5-large+BM25 $+$ GraphER-GCS                          & \textbf{11.4} (\textbf{26.0}) \\
          \rowcolor{gray!20}
          \textit{GraphER-GCS's improvement}                      & $+0.8$ ($+0.8$) \\
          \midrule
          \multicolumn{2}{c}{PR@10 (\%)} \\
          \midrule
          Llama-Embed-Nemotron-8B+BM25                                    & 71.3 (86.6) \\
          Llama-Embed-Nemotron-8B+BM25 $+$ GraphER-PPR    & 71.3 (86.1) \\
          Llama-Embed-Nemotron-8B+BM25 $+$ GraphER-GCS    & \textbf{73.4} (\textbf{87.9}) \\
          Llama-Embed-Nemotron-8B+BM25 $+$ GraphER-GAT    & \underline{73.0} (\underline{87.4}) \\
          \rowcolor{gray!20}
          \textit{GraphER-GAT's improvement}                      & $+1.7$ ($+0.7$) \\
          \midrule
          Multilingual-E5-large+BM25                                                          & \underline{17.0} (\underline{32.9}) \\
          Multilingual-E5-large+BM25 $+$ GraphER-PPR                          & 13.8 (29.1) \\
          Multilingual-E5-large+BM25 $+$ GraphER-GCS                          & \textbf{19.8} (\textbf{34.5}) \\
          \rowcolor{gray!20}
          \textit{GraphER-GCS's improvement}                      & $+2.9$ ($+1.6$) \\
          \bottomrule
        \end{tabular}
        }
    \end{scriptsize}
  \end{center}
  \caption{Retrieval performance on BEIR-NQ. Numbers in parentheses denote PR@K over all queries; numbers outside parentheses denote PR@K for queries with more than one relevant object.}
  \label{tab:chunked_retrieval_eval}
\end{table}

For the GAT rankers, we first train the neural network on \texttt{Bird\_train} while monitoring its performance on a sample of \texttt{Spider1\_train}. The validation results on this sample of \texttt{Spider1\_train} are used to select the learning rate, L2 regularization coefficient, and number of training epochs. After fixing these hyper-parameters, we retrain the network using the combined set of \texttt{Bird\_train} and \texttt{Spider1\_train}, and report their evaluation metrics on the \texttt{dev} and \texttt{test} splits of these two datasets, as well as on other datasets. The GAT rankers learn edge weights automatically via the attention mechanism, so users do not need to specify edge weights in this case.

\begin{table*}[!t]
  \begin{center}
    \begin{scriptsize}
        \resizebox{\linewidth}{!}{
        \begin{tabular}{lcccccccc}
          \toprule
                    & \multicolumn{8}{c}{\textbf{Dataset}} \\
          \cmidrule(lr){2-9}          
          \textbf{Reranker} & Spider1\_dev & Spider1\_test & Bird\_dev & Beaver\_dev & HotpotQA & 2WikiMultihopQA & MuSiQue & BEIR-NQ \\
          \midrule
          \multicolumn{9}{c}{PR@5 (\%)} \\
          \midrule
          GraphER-GCS                                  & 47.6 (61.0) & 29.2 (44.6) & 29.4 (36.6) & \textbf{17.0} (\textbf{17.7}) & 78.8 & 43.8 & 25.4 & 52.3 (78.1)  \\
          GraphER-GAT                                  & \textbf{49.7} (\textbf{62.5}) & \textbf{40.9} (\textbf{54.4}) & \textbf{36.9} (\textbf{43.2}) & 15.0 (15.8) & 78.0 & \textbf{44.1} & \textbf{25.6} & \textbf{54.4} (\textbf{78.2})  \\
          GraphER-MLP                                  & 45.2 (60.5) & 36.7 (52.7) & 35.2 (42.5) & 15.0 (15.8) & \textbf{78.9} & 42.5 & 21.6 & 50.3 (76.2)  \\
          \midrule
          \multicolumn{9}{c}{PR@10 (\%)} \\
          \midrule
          GraphER-GCS                                  & 69.0 (77.4) & 50.4 (64.2) & 72.8 (77.0) & \textbf{31.1} (\textbf{31.6}) & 88.7 & 51.1 & 36.9 & \textbf{73.4} (\textbf{87.9})  \\
          GraphER-GAT                                  & \textbf{70.1} (\textbf{78.5}) & \textbf{59.1} (\textbf{71.0}) & 77.0 (80.4) & 29.6 (30.1) & \textbf{88.9} & \textbf{53.0} & \textbf{37.4} & 73.0 (87.4)  \\
          GraphER-MLP                                  & 67.2 (76.0) & 55.1 (69.2) & \textbf{79.0} (\textbf{82.1}) & 28.2 (28.7) & 87.2 & 51.1 & 32.4 & 71.2 (85.6)  \\
          \bottomrule
        \end{tabular}
        }
    \end{scriptsize}
  \end{center}
  \caption{Comparison of GraphER-GCS, GraphER-GAT, and GraphER-MLP on evaluation benchmarks. The base retriever is Llama-Embed-Nemotron-8B+BM25.}
  \label{tab:message_passing_effect}
\end{table*}

\subsection{Retrieval Performance}
In Tables \ref{tab:table_retrieval_eval}, \ref{tab:multihop_retrieval_eval}, and \ref{tab:chunked_retrieval_eval}, we list the PR@5 and PR@10 for the three retrieval tasks. In terms of PR@10, the GraphER-GAT method improves upon the baseline in 14 out of 15 dataset-base retriever settings where it was applied. The GraphER-GCS method improves upon the baseline in all 18 dataset-base retriever settings where it was applied. These improvements are particularly pronounced for queries with more than one relevant object.

The PPR-variant of GraphER outperforms the baseline in 12 out of 18 dataset-base retriever settings. Moreover, in two of these settings, it leads to regressions of more than 7\% compared to the baseline. Our examination of its failure cases reveals that this occurs because hub objects are almost always ranked at the top under the PPR algorithm, regardless of their true relevance to the query.


\section{Discussions}
\label{sec:Discussions}

\subsection{Capturing Proximities Beyond Semantics}
Our experiments show that the improvements provided by GraphER-GCS and GraphER-GAT are robust to the choice of semantic embedding model used in the base retriever. This can be attributed to the fact that GraphER does not alter the object representations within the semantic space. Instead, it leverages proximity relationships that extend beyond semantic proximity. These additional proximity relationships reside in dimensions that are linearly independent of, or orthogonal to, the semantic space and are therefore not fully captured by similarity measures such as cosine or L2 distance between semantic embeddings.

GraphER incorporates these additional proximity relationships by explicitly adding edges between data objects that are proximal to each other. In this work, we identified and experimented with three commonly encountered proximity types beyond semantic proximity: structural proximity in table retrieval, conceptual proximity in multi-hop QA retrieval, and contextual proximity in chunked document retrieval. It is worth noting that these three types are not exhaustive, and the proximity relationships can be adapted to accommodate specific use cases or to inject domain-specific business logic.

\subsection{Learning Higher-Order Dependencies}

Both PPR and GCS are linear transformations of the initial retrieval scores. To account for higher-order interactions among node features, we additionally employ a graph attention network (GAT) as the ranking component within the GraphER framework. Across the evaluated settings, GraphER-GAT outperforms GraphER-GCS in terms of PR@10 in 13 out of the 15 dataset-base retriever settings in which it is applied.

To determine whether these gains arise from message passing rather than from additional nonlinear transformations, we conduct an ablation study in which the message-passing component of the GAT is disabled while the number of layers, hidden dimensions, and input features remain unchanged. This modification reduces the GAT to a per-node multilayer perceptron (MLP) applied independently to each node, which we refer to as GraphER-MLP. The MLP models are then trained using the same protocol. The comparison of its retrieval performance to GraphER-GCS and GraphER-GAT is reported in Table \ref{tab:message_passing_effect}.

GraphER-GAT achieves the best PR@10 on 5 of the 8 datasets, whereas GraphER-MLP achieves the best result on only one. These results suggest that GraphER-GAT benefits from message passing across graph edges, rather than relying solely on per-node feature transformations.

\subsection{Retrieval Latency}

GraphER involves the online construction of an $n \times n$ adjacency matrix, where $n$ is the number of candidate objects. The additional time and space complexity introduced by GraphER beyond the base retriever can be controlled by adjusting this parameter. When GraphER-GAT is used, the GAT model incurs an additional forward pass. 

Our experiments were conducted on a compute node with two AMD EPYC 7J13 processors, 1 TB of RAM, and a single NVIDIA A100-SXM4-40GB GPU. Empirically, GraphER-GCS and GraphER-GAT required 0.49 and 0.55 seconds, respectively, to rerank 200 candidates, which is negligible compared with downstream LLM inference latency.

\subsection{Compatibility with Semantic Rerankers}

GraphER is not intended to replace neural rerankers such as cross-encoder models. Instead, it addresses a complementary problem. Cross-encoder rerankers refine semantic relevance by jointly encoding each query-document pair, whereas GraphER exploits relationships among candidate objects that are not captured by the point-wise semantic scoring. Because GraphER is computationally lightweight and operates independently of the underlying semantic scorer, it can be combined with cross-encoder rerankers in a multi-stage retrieval pipeline. Evaluating such hybrid pipelines is an important direction for future work.

\subsection{Limitations and Future Work}

GraphER relies on the availability and quality of metadata that encode proximity relationships among data objects. When such relationships are noisy, incomplete, or weakly aligned with the information need, graph-based reranking may provide limited benefits to retrieval performance. Furthermore, although the GraphER framework is designed to support arbitrary forms of proximity, our experiments cover only three representative types: structural, conceptual, and contextual proximity. Additional studies are needed to evaluate the effectiveness of other enrichment strategies across different domains. Finally, like other reranking approaches, GraphER operates only on the candidate set returned by the base retriever and therefore cannot recover relevant objects that are absent from the initial retrieval results. Consequently, its effectiveness depends on the base retriever achieving sufficient recall during candidate retrieval.

Several directions remain for future work. First, although we demonstrate that GraphER can be combined with graph neural networks, exploring more expressive graph learning architectures and alternative graph construction strategies may further improve retrieval quality. Second, while this work evaluates GraphER as a standalone reranking component to isolate its contribution, it is also compatible with downstream semantic rerankers, such as cross-encoder models. An empirical evaluation of such hybrid pipelines would help quantify the complementary benefits of improving candidate completeness and fine-grained semantic relevance. Finally, GraphER currently uses user-defined or extracted proximity relationships. Learning proximity relationships automatically from retrieval or user interaction data may enable the framework to adapt more effectively to different domains and applications.

\section{Conclusion}
\label{sec:Conclusion}

We presented GraphER, a graph-based enrichment and reranking framework for improving retrieval completeness in RAG systems. GraphER incorporates proximity relationships beyond semantic similarity through lightweight metadata enrichment and query-time graph construction, without requiring additional graph infrastructure. Across table retrieval, multi-hop QA retrieval, and chunked-document retrieval tasks, GraphER consistently improves retrieval completeness over strong semantic retriever baselines. We also introduced Graph Cohesive Smoothing, which provides more robust gains than Personalized PageRank, and showed that a graph attention network can further improve performance by modeling higher-order interactions among candidate objects. These results demonstrate that leveraging organizational structure beyond semantic similarity is an effective and practical approach for improving retrieval in RAG systems.

\bibliography{custom}

\onecolumn

\appendix
\raggedbottom

\section{Appendix}
\label{sec:appendix}

\subsection{Prompt Template for Named Entity Extraction}
\label{appendix:prompt_ner}

\begin{tcolorbox}[
  colback=white,
  colframe=black,
  boxrule=1.2pt,
  arc=8pt,
  left=10pt,
  right=10pt,
  top=10pt,
  bottom=10pt,
  before=\vspace{3pt},
  after=\vspace{3pt},
]
\begin{MyVerbatim}
Identify the key entities (people, organizations, locations, dates, etc.) from the following paragraph.
Return the result as a valid Python list of strings, with no explanations, only the list.

Example:
Text:
"Barack Obama was born in Honolulu, Hawaii, and served as the 44th President of the United States."
Output:
["Barack Obama", "Honolulu", "Hawaii", "United States", "44th President"]

Text:
{content}
Output:
\end{MyVerbatim}
\end{tcolorbox}

\subsection{Prompt Template for Retrieval-Augmented Answer Generation}
\label{appendix:prompt_rag}

\begin{tcolorbox}[
  colback=white,
  colframe=black,
  boxrule=1.2pt,
  arc=8pt,
  left=10pt,
  right=10pt,
  top=10pt,
  bottom=10pt,
  before=\vspace{3pt},
  after=\vspace{3pt},
]
\begin{MyVerbatim}
You are an AI assistant. Use ONLY the information provided in the context below to answer the user's question.

If the context does not contain enough information to answer the question, say:
"I don't have enough information in the provided context to answer this."

<context>
{retrieved_objects}
</context>

Question:
{query}

Answer:
\end{MyVerbatim}
\end{tcolorbox}

\subsection{Prompt Template for RAG Answer Evaluation}
\label{appendix:prompt_rag_eval}

\begin{tcolorbox}[
  colback=white,
  colframe=black,
  boxrule=1.2pt,
  arc=8pt,
  left=10pt,
  right=10pt,
  top=10pt,
  bottom=10pt,
  before=\vspace{3pt},
  after=\vspace{3pt},
]
\begin{MyVerbatim}
You are an impartial evaluator tasked with judging the correctness of an answer generated by an LLM.

=== Task ===
Evaluate whether the Ground Truth Answer is present in the Generated Answer.

Your goal is to determine if the essential meaning and key facts of the Ground Truth Answer are correctly captured in the Generated Answer.

=== Instructions ===
1. Carefully compare the Generated Answer with the Ground Truth Answer.
2. Focus on substance and meaning, not exact wording.
3. Treat paraphrases, synonyms, or logically equivalent statements as correct if they preserve the same factual meaning.
4. Ignore irrelevant extra information unless it directly contradicts the Ground Truth.
5. If any vital fact from the Ground Truth is missing or incorrect, the decision should be "no".
6. If all essential information from the Ground Truth is present and correct, the decision should be "yes".

=== Input Data ===
- Question: {query}
- Generated Answer: {generated_answer}
- Ground Truth Answer: {ground_truth_answer}

=== Output Format ===
Provide your final evaluation strictly in the following format:

Explanation: <Brief explanation of how you made the decision>
Decision: <yes or no>

Proceed carefully and base your judgment strictly on the criteria above.
\end{MyVerbatim}
\end{tcolorbox}

\end{document}